\documentclass{article}
\pdfoutput=1
\usepackage{amsmath} 
\usepackage{PRIMEarxiv}
\usepackage[mathscr]{euscript}

\usepackage{bm}
\usepackage{amssymb}
\usepackage{array}
\usepackage[utf8]{inputenc} 
\usepackage[T1]{fontenc}    
\usepackage{hyperref}       
\usepackage{url}            
\usepackage{booktabs}       
\usepackage{amsfonts}       
\usepackage{nicefrac}       
\usepackage{microtype}      
\usepackage{lipsum}
\usepackage{fancyhdr}       
\usepackage{graphicx}       
\graphicspath{{media/}}     

\title{Adaptive Depth Graph Attention Networks
}

\author{
  Jingbo Zhou , Yixuan Du , Ruqiong Zhang , Rui Zhang{*}\\
Key Laboratory of Symbolic Computation and Knowledge Engineering of MOE, Jilin University, China\\
  \texttt{\{zhoujb5520,duyx5520,zhangrq1320\}@mails.jlu.edu.cn,rui@jlu.edu.cn}\\
}

\begin{document}
\maketitle

\begin{abstract}
As one of the most popular GNN architectures, the graph attention networks (GAT) is considered the most advanced learning architecture for graph representation and has been widely used in various graph mining tasks with impressive results. However, since GAT was proposed, none of the existing studies have provided systematic insight into the relationship between the performance of GAT and the number of layers, which is a critical issue in guiding model performance improvement. In this paper, we perform a systematic experimental evaluation and based on the experimental results, we find two important facts: (1) the main factor limiting the accuracy of the GAT model as the number of layers increases is the oversquashing phenomenon; (2) among the previous improvements applied to the GNN model, only the residual connection can significantly improve the GAT model performance. We combine these two important findings to provide a theoretical explanation that it is the residual connection that mitigates the loss of original feature information due to oversquashing  and thus improves the deep GAT model performance. This provides empirical insights and guidelines for researchers to design the GAT variant model with appropriate depth and well performance. To demonstrate the effectiveness of our proposed guidelines, we propose a GAT variant model-ADGAT that adaptively selects the number of layers based on the sparsity of the graph, and experimentally demonstrate that the effectiveness of our model is significantly improved over the original GAT.
\end{abstract}


\section{INTRODUCTION}
In recent years, GNNs\cite{inproceedings,DBLP:journals/tnn/ScarselliGTHM09,DBLP:journals/tnn/Micheli09} have become the state-of-the-art model for processing graph-structured data and can be widely used in various fields such as social networking\cite{DBLP:journals/corr/abs-1902-07243,DBLP:journals/corr/abs-2110-03987}, product recommendation\cite{DBLP:journals/corr/MontiBB17,DBLP:journals/csur/WuSZXC23,DBLP:journals/kbs/YinLZL19} and drug discovery\cite{DBLP:journals/corr/abs-2102-10056,DBLP:journals/jcheminf/JiangWHCLWSCWH21}, etc. GNNs provide a general framework for learning from graph-structured data through a message-passing mechanism that aggregates representations of neighboring nodes. 

As GNN models are further used and studied, it is found that deeper models are often needed in many cases to obtain better results. For example, Wentao Zhang et al.\cite{DBLP:journals/corr/abs-2108-00955} illustrates the necessity of deep GNN layers from the perspective that sparse graphs, which are more common in reality, require larger receptive field; Uri Alon and Eran Yahav \cite{DBLP:journals/corr/abs-2006-05205} shows the need to utilize deeper GNN layers also in some scenarios where it should be used and updating the node representation requires distant nodes. However, as the number of model layers increases, the effectiveness of the GNN model often decreases significantly, and thus does not reach the required depth. For the problem that GNN cannot be deepened, many articles have proposed effective methods.

Previous work has mainly focused on the study of deepening techniques for GCN-style models without attention mechanisms\cite{DBLP:conf/icml/XuLTSKJ18,DBLP:conf/iccv/Li0TG19,inproceedings1,DBLP:journals/corr/abs-1902-07153,DBLP:journals/corr/abs-2004-11198}, while little consideration has been given to GNN models with attention mechanisms, such as GAT. As one of the most popular GNN architectures, Graph Attention Networks (GAT) proposed by Velickoviˇc et al.\cite{DBLP:journals/corr/abs-1710-10903} is considered the most advanced graph representation learning architecture. Compared with the previous varieties of other GCN and its variants based on spectral graph theory, GAT references the self-attention mechanism in the transformer model, enabling nodes to calculate the importance of neighbor node representations to themselves in advance and perform neighborhood aggregation based on the learned importance coefficients. Considering the differences between the two architectures of GCN and GAT, the improvement measures for deep GNN in previous work may not fit well into the GAT model. To address the effectiveness of GAT when deepening the layers, some related work has emerged recently. G Dasoulas et al.\cite{DBLP:journals/corr/abs-2103-04886} proposed a Lipschitz normalization technique to enhance the Lipschitz continuity of the model to solve the gradient explosion problem of the model; Guangxin Su et al.\cite{su2023simple} proposed LSGAT based on a new attentional computation to solve the oversmoothing problem. Although good progress has been made in the current work, GAT still lacks as comprehensive and in-depth studies as GCN for layer-related problems. Previous studies have speculated on the various types of problems that GNNs may encounter when deepening, such as overfitting\cite{DBLP:journals/corr/abs-1907-10903}, overcorrelation\cite{10.1145/3534678.3539445}, oversmoothing\cite{DBLP:journals/corr/abs-2010-02863,Godwin2021VeryDG,DBLP:journals/corr/abs-2003-08414,DBLP:journals/corr/abs-2102-06462,DBLP:journals/corr/abs-1909-12223}, etc. Which are the main problems that affect GAT performance as the depth of GAT increases? What previous improvements applied to GNN models can alleviate this problem? Based on this, we focus on various issues related to the depth of GAT.

Our work can be summarized as follows:

1. We refer to the factors that were considered to cause possible degradation of GNN‘s effectiveness in the past, such as overcorrelation, oversmoothing, oversquashing, gradient vanishing, etc., and experimentally verify that many existing studies that suggest possible problems with deeper GAT actually do not exist. We also find that the main problem that affects the performance of GAT as the number of layers increases is the oversquashing phenomenon that occurs when aggregating messages across long paths.

2. We investigate operation and components previously applied to GNNs models, and evaluate their effectiveness in mitigating the performance problem when the GAT model is deepened. After that, we combined the experimental results to propose a theoretical explanation: as the GAT deepens, the oversquashing phenomenon is aggravated and the loss of the original feature information of the nodes occurs.

3. Based on our findings, we propose a GAT variant model-ADGAT, and experimentally validate that the model's effectiveness has more significant improvement using  initial residual connections and the mechanism of adaptive selection of network layers for graphs with different sparsity.

\section{PRELIMINARY}
\label{sec:headings}

Consider a undirected graphs with N nodes denoted by $\mathscr{G}( \mathscr{V},\mathscr{E},\mathscr{X})$, where  $\mathscr{V} = \{1,2 ...n\}$ denotes the set of N nodes, $\mathscr{E} \subset \mathscr{V}\times\mathscr{V}$ denotes the set of the edge and $\mathscr{X} = \in \mathbb{R}^{N\times d}$ denotes the node feature matrix, where d indicate the dimension of node feature.Let $X_i$ denotes the i-th row of the feature matrix $\mathscr{X}$ and the representation of node $v_i$. let A be the adjacency matrix, where $A_{i,j} = 1$ if $(i,j)\in \mathscr{E}$ , otherwise $A_{i,j} = 0$ .let D be the diagonal degree matrix

\subsection{Graph Neural Networks(GNN) }

We define the Graph Neural Networks (GNN)  as a model that utilizes the message passing mechanism to aggregate neighbor nodes to update the representation of each node. For each GNN layer, let $\left\{\boldsymbol{h}_i \in \mathbb{R}^d \mid i \in \mathcal{V}\right\}$, a set of node representation, be the input and$\left\{\boldsymbol{h}_i^{\prime} \in \mathbb{R}^{d^{\prime}} \mid i \in \mathcal{V}\right\}$ be the output, then the operation of the GNN layer can be defined as

\begin{equation}
\boldsymbol{x}_{\boldsymbol{i}}^{\prime}=g_\theta\left(\boldsymbol{x}_i, \text { AGGREGATE }\left(\left\{\boldsymbol{x}_j \mid j \in \mathcal{N}_i\right\}\right)\right)
\end{equation}

where AGGREGATE and g represent aggregation function and update function respectively, and g is parameterized by $\theta$. Different GNN models mainly differ from the different designs of these two functions. In this paper, when we refer to the proposition that "GNN performance decreases with increasing depth", we mainly refer to Graph Convolution Network(GCN).

\subsection{Graph Convolution Network(GCN)}
GCN is the most important  GNN model, which is essentially a first-order approximation of Chebynet\cite{DBLP:conf/nips/DefferrardBV16}. The operation of the GCN layers can be implemented with the following formula:

\begin{equation}
\mathrm{X}^{(l)}=\sigma\left(\tilde{\mathrm{D}}^{-1 / 2} \tilde{\mathrm{A}} \tilde{\mathrm{D}}^{-1 / 2} \mathrm{X}^{(l-1)} \mathrm{W}^{(l)}\right)
\end{equation}

where $\tilde{\mathrm{D}} $ denotes the diagonal 
degree matrix of $\tilde{\mathrm{A}}$ and $\tilde{\mathrm{A}} = I + A $. $\mathrm{W}^{(l)}$ is a learnable parameter matrix. $\sigma$ denotes the activation function.

\subsection{Graph Attention Network(GAT)}
Different from GCN and many other popular GNN architectures Some models\cite{DBLP:journals/corr/HamiltonYL17,DBLP:journals/corr/abs-2006-13318} that weight all neighbors with equal importance, GAT draws on the self-attention mechanism in the transformer to allow nodes to calculate the importance of its neighbors $j \in \mathcal{N}_i$, and use the obtained attention scores for weighted summation during aggregation. 
The operation of the GAT layers can be implemented by the following formula:

\begin{equation}
e\left(\boldsymbol{h}_i, \boldsymbol{h}_j\right)=\operatorname{LeakyReLU}\left(\boldsymbol{a}^{\top} \cdot\left[\boldsymbol{W} \boldsymbol{h}_i \| \boldsymbol{W} \boldsymbol{h}_j\right]\right) \label{e3}
\end{equation}

\begin{equation}
\alpha_{i j}=\operatorname{softmax}_j\left(e\left(\boldsymbol{h}_i, \boldsymbol{h}_j\right)\right) 
=\frac{\exp \left(e\left(\boldsymbol{h}_i, \boldsymbol{h}_j\right)\right)}{\sum_{v_{j^{\prime}} \in \mathcal{N}_i} \exp \left(e\left(\boldsymbol{h}_i, \boldsymbol{h}_{j^{\prime}}\right)\right)}
\label{e4}
\end{equation}

\begin{equation}
\boldsymbol{h}_i^{\prime}=\sigma\left(\sum_{j \in \mathcal{N}_i} \alpha_{i j} \cdot \boldsymbol{W} \boldsymbol{h}_j\right) \label{e5}
\end{equation}

where $\boldsymbol{h}_i$ denotes the representation of the node i.
Eq.\ref{e3} calculates the attention score between a pair of nodes i and j, where $
\boldsymbol{a} \in \mathbb{R}^{2 d^{\prime}} , \boldsymbol{W} \in \mathbb{R}^{d^{\prime} \times d}
$ are learnable parameters, $\|$ represents the concat operation, and LeakyRelu is an activation function. Eq.\ref{e4} normalizes the attention scores of node i and its neighbors $j \in \mathcal{N}_i$ obtained by Equation 3. Eq.\ref{e5} performs a weighted summation of the neighbor node representations according to the normalized attention scores.

\subsection{EP and ET Operation}
Wentao Zhang et al.\cite{DBLP:journals/corr/abs-2108-00955} decoupling analysis of the convolution operation of GNN, each graph convolutional layer in GNN can be divided into two operation: Embedding Propagation (EP) and Embedding Transformation (ET). In the GCN-style model, the EP operation is defined as:

\begin{equation}
    \mathrm{EP}(\mathrm{X})=\hat{\mathrm{A}} X,
\end{equation}

ET operation is defined as: 

\begin{equation}
\operatorname{ET}(\hat{\mathrm{X}})=\sigma(\hat{\mathrm{X}} \mathrm{W}),
\end{equation}

However, due to the difference in GCN and GAT architectures, we define Eq. \ref{e3}, \ref{e4} in the GAT framework as EP operation, and Eq.\ref{e5} as ET operation.

\textbf{Depth of EP and ET operation} In this paper, we define the number of EP and ET operation performed in the model as the depth of EP and ET operation, denoted by $D_p$ and $D_t$, respectively.

\section{MISCONCEPTIONS}

\subsection{Experimental Setup}
The Pubmed dataset includes 19717 scientific publications on diabetes from the Pubmed database, and the citation network consists of 44338 edges. Among the three major underlying graph network datasets (Pubmed\cite{Sen2008CollectiveCI}, Cora\cite{Sen2008CollectiveCI}, and Citeseer\cite{Sen2008CollectiveCI}), considering Pubmed as the dataset with the most nodes and edges, we believe it has a greater advantage in terms of accuracy measures of the model and credibility of the experimental results. Therefore, we default to perform the node classification task on Pubmed for various types of exploration experiments. In addition, we use the Cora dataset for the activation function and the FA layer experiments.

To make the experimental results more reliable, ten random seeds were randomly selected for each experiment and the mean value of the experimental results was taken as the final result.

\subsection{Overfitting}
Some studies\cite{DBLP:journals/corr/abs-1907-10903,DBLP:journals/corr/abs-2003-13663,DBLP:journals/corr/abs-1904-03751,DBLP:journals/corr/abs-2006-07107} have attributed the degraded performance of deep GNNs to overfitting. The overfitting phenomenon refers to the result that the model is incapable of fitting the rest of the data well due to overly close or exact matching of a limited training data set, which in turn results in an excellent performance on the training set, but poor performance on the test set.

To investigate the overfitting phenomenon in GAT models with different depths, we conducted node classification experiments using GAT models with depths of 2, 5, 10, 15, 20, 25, and 30 to compare the prediction accuracy of these models on the training and test sets.

\begin{figure}[htbp]
\includegraphics[scale=0.6]{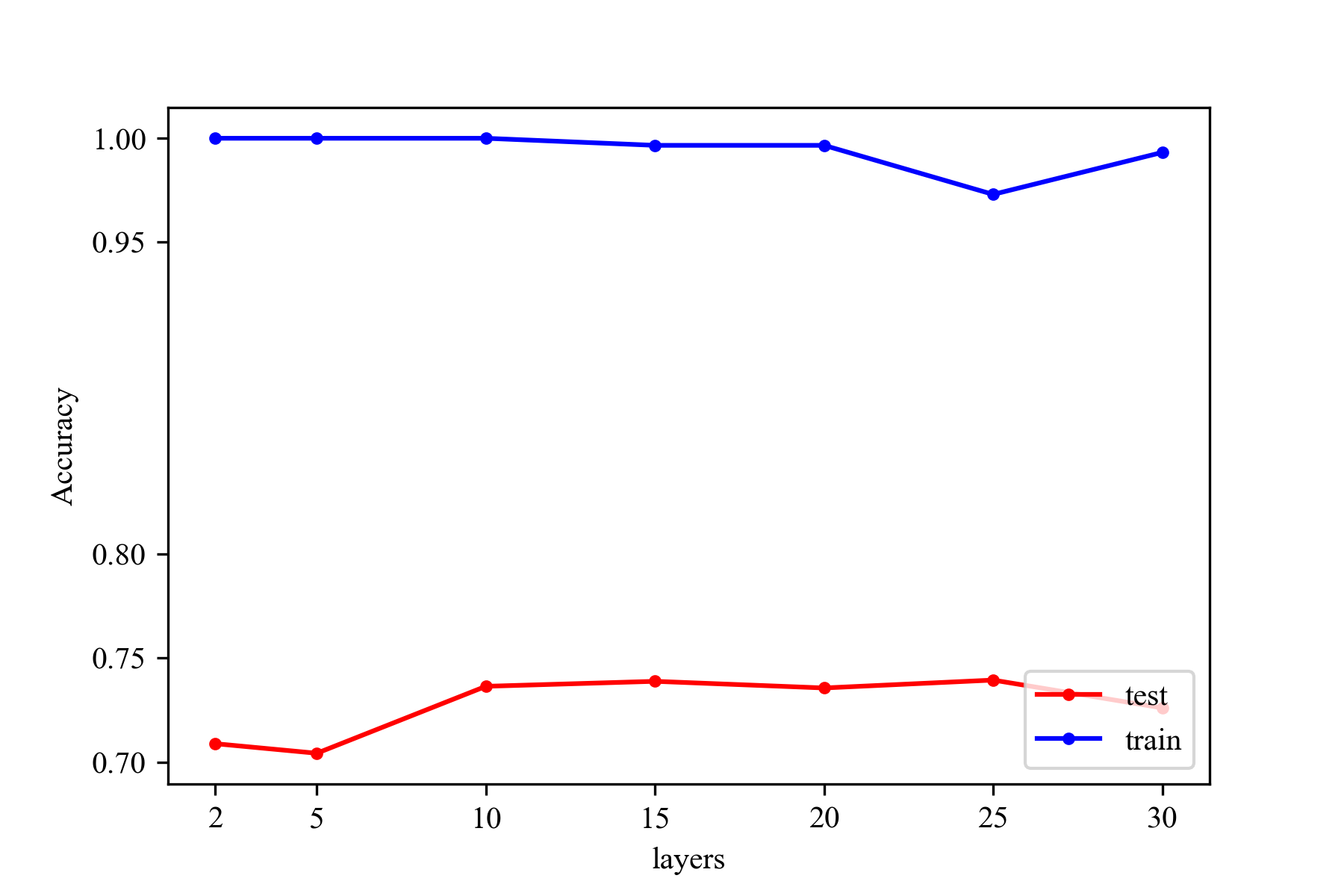}
\label{fig01}
\centering
\caption{Experimental result of overfitting.}
\end{figure}

The experimental results are shown in Figure \ref{fig01}. We find that the prediction accuracy of the GAT model on the training and test sets does not show overfitting as the number of layers increases. It indicates that deepening the number of GAT layers does not lead to overfitting and thus a decrease in the effect, but at the same time the effect is not improved, so overfitting is not the main factor affecting the model performance when GAT is deepened.

\subsection{Oversmoothing}
Previous work\cite{DBLP:journals/corr/abs-1801-07606,DBLP:journals/corr/abs-1906-01210} has shown that oversmoothing exists in GNNs. Oversmoothing means that as the number of operation to aggregate neighbor information increases, the embeddings of some nodes will converge to the same value with the features of different nodes tending to be homogeneous. Most previous studies have attributed the performance degradation of deep GNNs to oversmoothing. In contrast, many recent studies, such as the work of Wentao Zhang, Zeang Sheng\cite{DBLP:journals/corr/abs-2108-00955}, have discussed in more depth whether oversmoothing is the main factor limiting the number of GNN layers.

However, most of the above work has focused on GCN-related models and has not concentrated on the oversmoothing phenomenon in GAT models with the introduction of the self-attention mechanism. To explore the oversmoothing phenomenon in GAT, we conduct experiments to observe the changes in model accuracy and smoothness as the depth of GAT increases. Among them, we use the metric SMV, a measure of over-smoothing proposed by Liu et al.\cite{10.1145/3534678.3539445} to measure the smoothness of the current graph. Specifically, SMV normalizes the node representations and calculates their Euclidean distances as the following equation:

\begin{equation}
    D(x,y) = \frac{1}{2} \left \|{ \frac{x}{\Vert x \Vert} - \frac{y}{\Vert y \Vert}} \right \|_2
\end{equation}

\begin{equation}
    SMV(X) = \frac{1}{N(N-1)}\sum_{i\neq j}D(X_{i,:},X_{j,:})
\end{equation}

where $X_{i,:}$ denotes the i-th row of feature matrix X, which is the representation of node $v_i$,$D(x,y$ represents the normalized Euclidean distance. According to the definition of SMV, higher SMV values indicate lower smoothing. In other words, the more homogeneous the features tend to be.

\begin{figure}[htbp]
\includegraphics[scale=0.6]{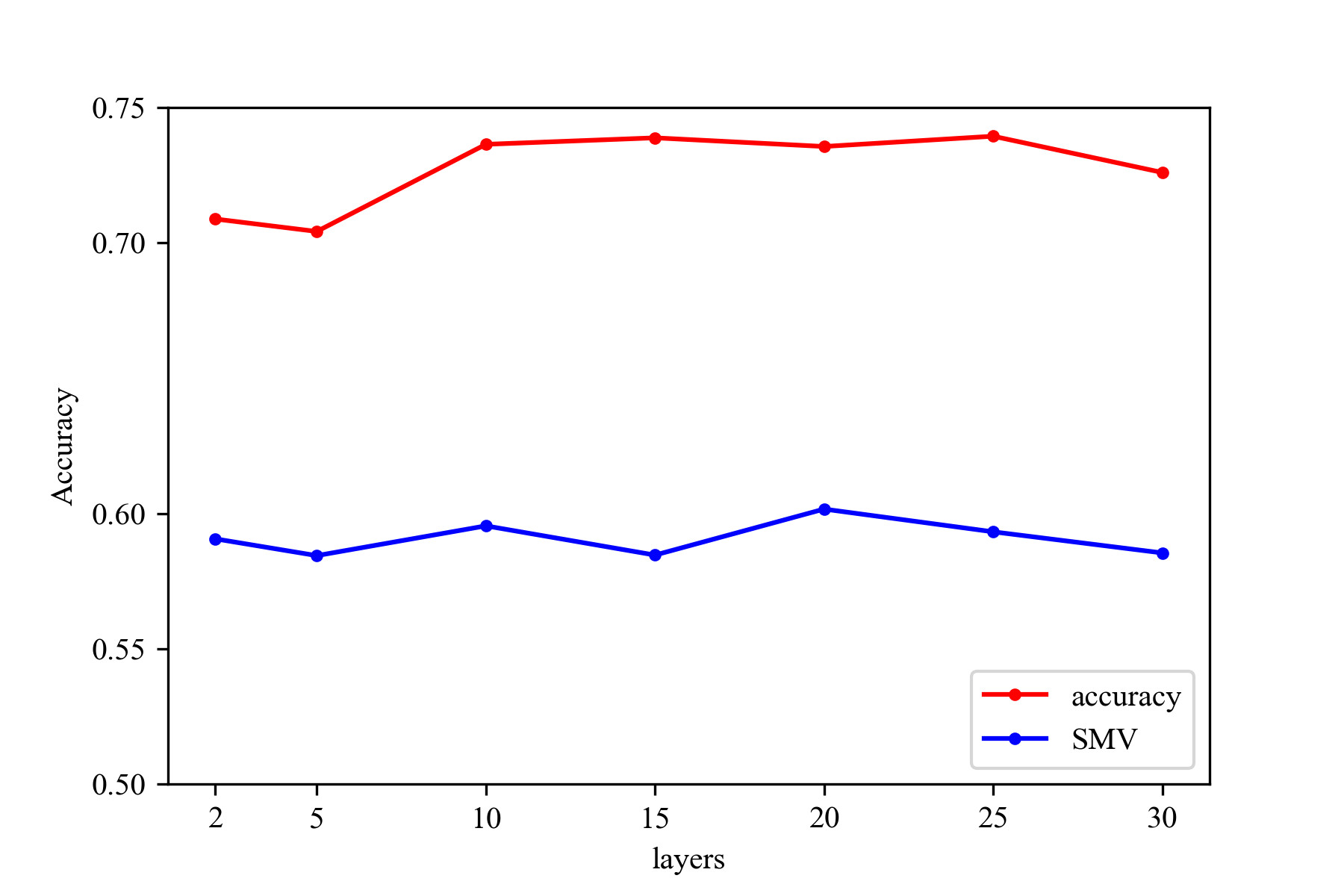}
\label{fig2}
\centering
\caption{Experimental result of oversmoothing.}
\end{figure}

We conducted node classification experiments on the Pubmed dataset using GAT models with depths of 2, 5, 10, 15, 20, 25, and 30, respectively. The observed variation of SMV values with the number of model layers is shown in Figure \ref{fig2}.

As the number of layers increases, the SMV value does not show a downward trend but oscillates steadily around 0.59, there is also no significant correlation between model accuracy and SMV values. From this view, the deep GAT did not show any over-smoothing, which means further that the test accuracy change of GAT is not affected by the oversmoothing problem when the number of layers increases. It is slightly different from the results in the literature \cite{10.1145/3534678.3539445}, and we speculate that the difference in the results may come from the degree of initialization of the parameters before training.

\subsection{Overcorrelation}

Wei Jin et al.\cite{10.1145/3534678.3539445} found that when there are too many GNN layers, the final obtained node features have a high correlation between dimensions. This phenomenon is defined as overcorrelation, which implies high redundancy and less information encoded in the learned dimensions, thus compromising the downstream task effectiveness. This work suggested that the over-correlation phenomenon is the main factor that limits the number of GNN layers.

Inspired by that work, we think it is necessary to explore whether the GAT model accuracy is affected by the overcorrelation problem when the number of layers increases. We refer to the overcorrelation metric of that work for a comparison of overcorrelation in GAT models with different layers. The metrics are formulated as follows:

\begin{equation}
    Corr(X) = \frac{1}{d(d-1)}\sum_{i\neq j}\lvert p(X_{:,i},X_{:,j} \lvert  \qquad i,j \in[1,2,...,d]
\end{equation}

\begin{figure}[htbp]
    \centering
    \includegraphics[scale=0.6]{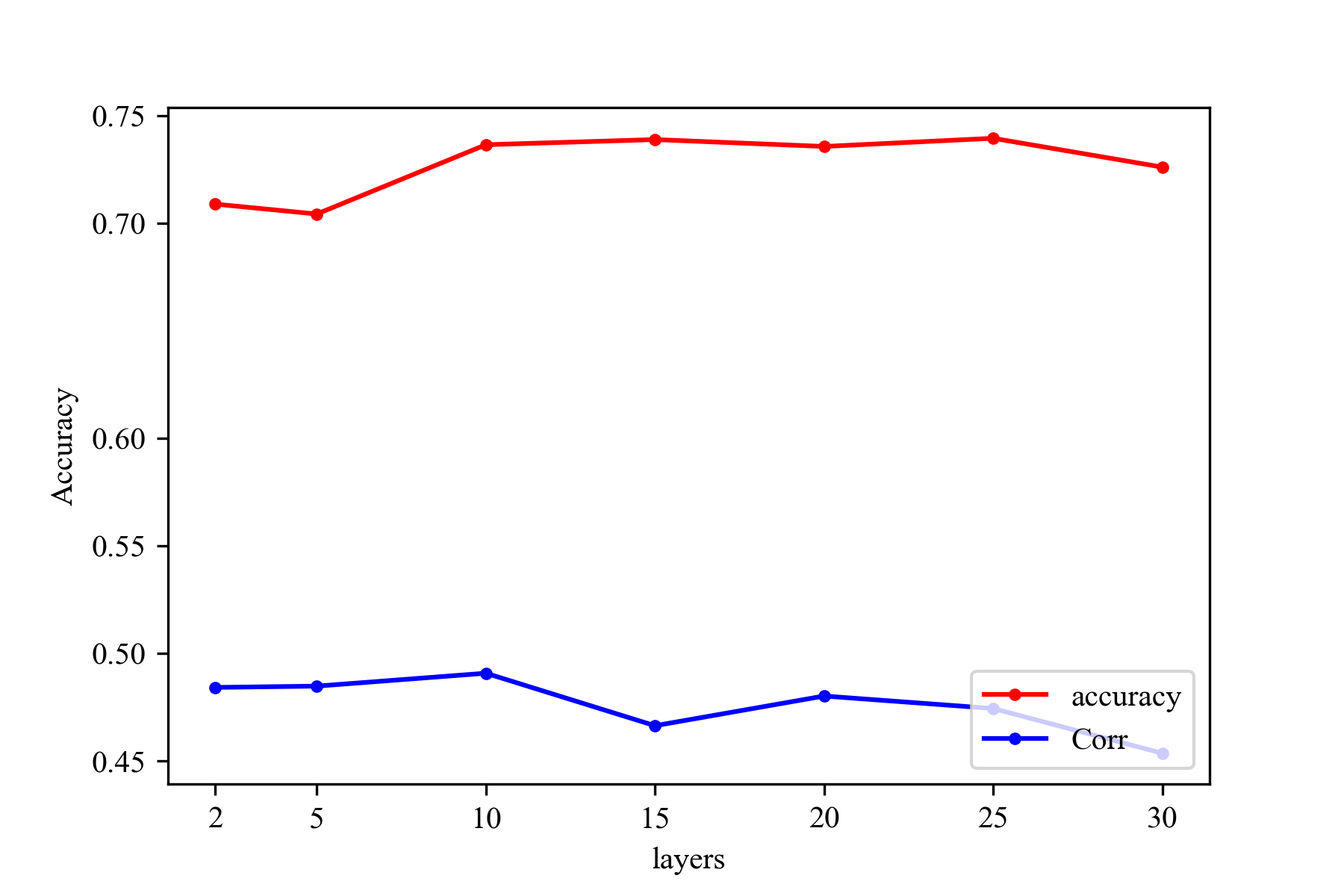}
    \caption{Experimental result of overcorrelation.}
    \label{fig3}
\end{figure}

We plot the accuracy of the GAT model with different numbers of layers on the Pubmed dataset as well as the Corr values in Figure \ref{fig3}.

Observed that the degree of overcorrelation does not fluctuate substantially when the number of GAT layers deepens, and the model accuracy does not change in the same magnitude as the fluctuation of the value of the overcorrelation metric. Therefore, we conclude that the accuracy of the deep GAT model is not affected by the overcorrelation problem.

Observed that the degree of over-correlation does not fluctuate substantially when the GAT layer is deepened, and the model accuracy does not change in the same magnitude as the fluctuation of the value of the over-correlation metric. Therefore, we conclude that the accuracy of the deep GAT model is not affected by the over-correlation problem.

\subsection{Gradient Vanishing}
Deep neural networks use the backpropagation method, which uses chain derivation and involves some concatenation operation when calculating the gradient of each layer. Therefore, as the depth of the network deepens, if most of the factors of concatenation are less than 1, the gradient update information will decay exponentially, and the final product may tend to 0, i.e. gradient vanishing, resulting in no change in the parameters of the subsequent layers of the network.

\begin{figure}[htbp]
    \centering
    \includegraphics[scale=0.6]{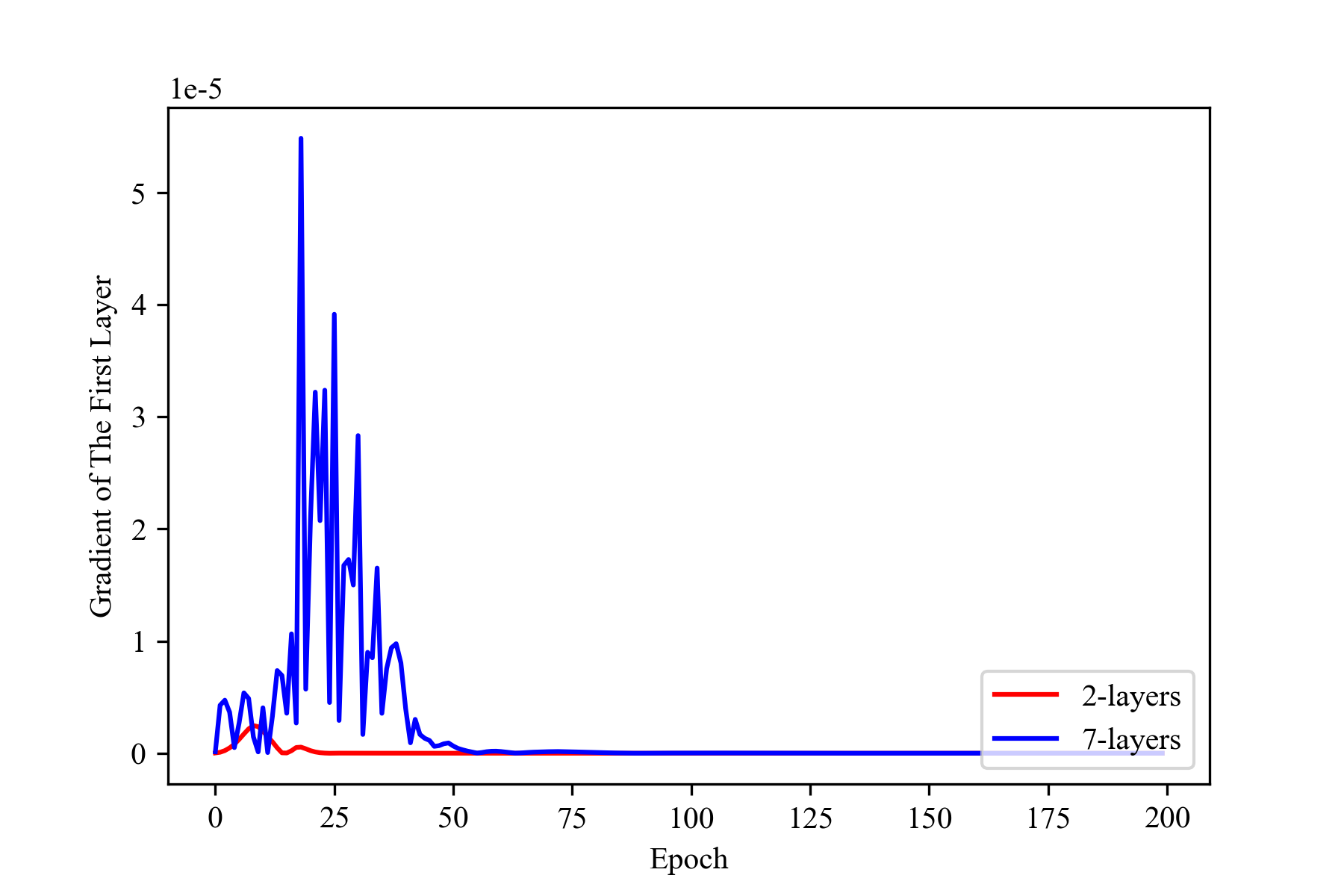}
    \caption{Experimental result of gradient vanishing.}
    \label{fig4}
\end{figure}

As the number of layers increases, GAT may also have the problem of gradient vanishing. To evaluate whether gradient vanishing occurs in deep GAT, we conduct node classification experiments on the Cora dataset and plot the variation of gradients during GAT training with different layers. The Figure \ref{fig4} shows the average absolute value of the first-layer gradient matrix of the 2-layer and 7-layer GAT during training.

The gradient of the 7-layer GAT fluctuates more at the beginning, which may be interpreted as the large models need more momentum to adjust in the early stage of training, and thus jump out of the suboptimal local minimum. And after the fluctuation, the gradients of both 7-layer GAT and 2-layer GAT tend to a stable low value. By looking at the training data, we can see that the accuracy of the training set has converged to be complete at this time, so the gradient is low. It means that compared with the shallow GAT, the deep GAT does not have the problem of gradient vanishing.

\subsection{Oversquashing}
The work of Uri AlonEran, Yahav et al.\cite{DBLP:journals/corr/abs-2006-05205} shows that GNNs are prone to the problem of oversquashing when aggregating messages across long paths. Tasks that rely on long-range interactions require more GNN layers, which results in an exponentially growing amount of information compressed into a fixed-length vector. As a result, GNNs are unable to propagate long-range information and can only learn short-range signals from training data, which makes GNNs perform poorly when the prediction task relies on long-range interactions. The more layers there are, the more severe the oversquashing becomes.

In order to study the impact of the squeeze problem on GAT, we increase the number of graph convolutional layers in GAT, set the number of neurons in each hidden layer to be twice that of the previous layer, and compare with the GAT with a constant number of hidden layer neurons. In contrast, the two models perform node classification tasks simultaneously.

\begin{figure}[htbp]
    \centering
    \includegraphics[scale = 0.6]{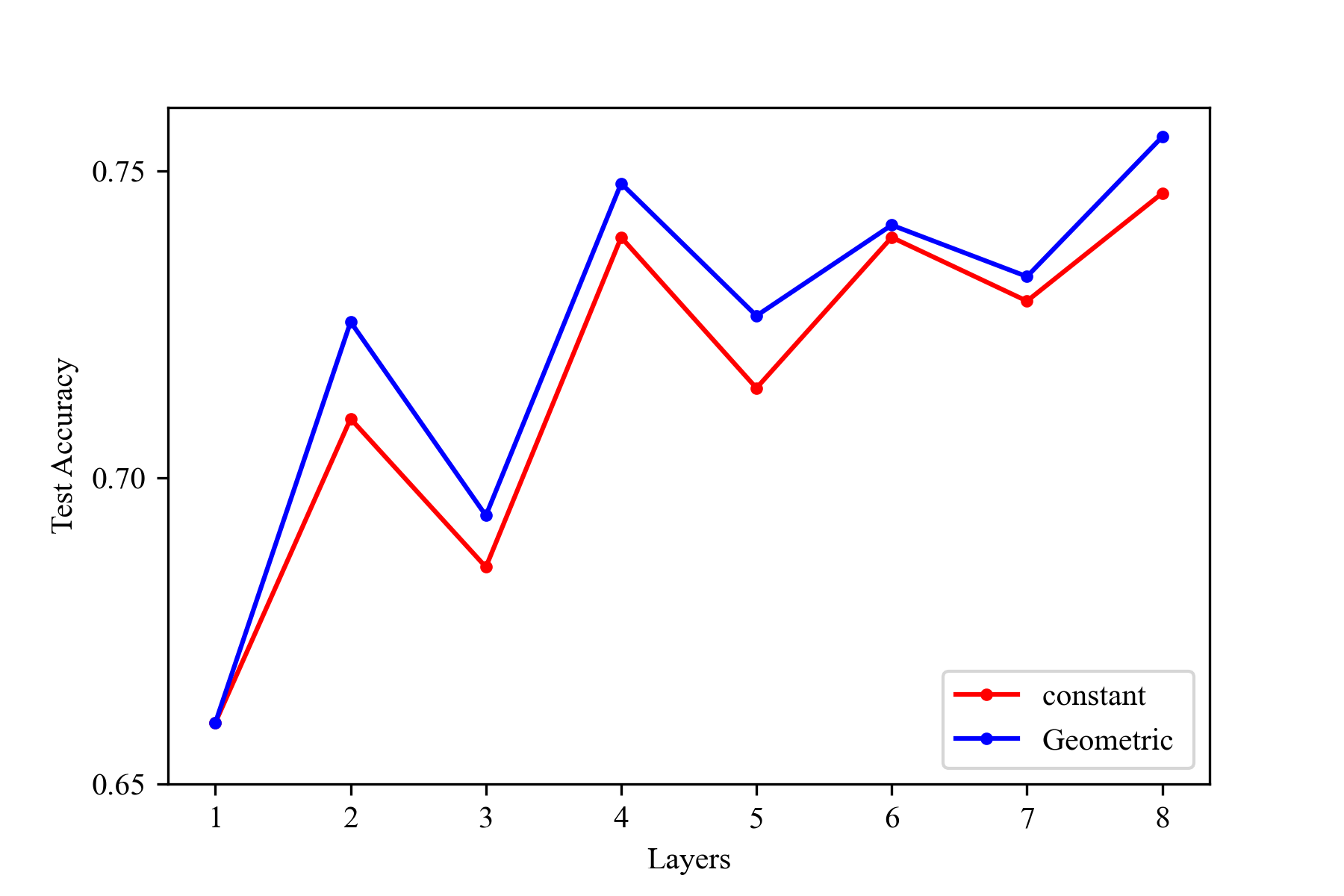}
    \caption{Experimental result of oversquashing.}
    \label{fig5}
\end{figure}

The results are shown in the figure \ref{fig5}. When the receptive field increases, the length of the vector can obviously hold more information, and when the number of layers increases, the performance of GAT with exponentially increasing number of neurons in the hidden layer is significantly improved compared with the original GAT.

We infer that the over-squashing problem is sufficiently improved by exponentially increasing the number of hidden layer neurons. Observing that the Pubmed dataset contains a total of 19717 nodes and 44338 undirected edges, each node in the graph is connected with about 4 edges on average, i.e., as the convolutional layer deepens, the receptive field increases and the amount of information grows exponentially, so that the vector dimension accommodating the information increases exponentially in parallel, which can alleviate the problem of over-squashing information compression.

In summary, the performance of GAT can improve significantly when the over-squashing problem is alleviated, i.e., the over-squashing problem is the main factor affecting the performance of GAT when its depth increases.

\subsection{Summary}
Although many works have successfully explored deep GNN architectures, what limits the performance of GAT models as the number of layers increases is not well understood systematically. Referring to the existing research, we analyzed the five possible causes: overfitting, oversmoothing, overcorrelation, gradient disappearance and oversquashing through systematic experiments, and found that the problems that many studies thought might occur as the number of GAT layers increases do not exist, such as overfitting, oversmoothing, overcorrelation, and gradient vanishing. Based on the experimental results, we find the most important factor that affects the performance of GAT as the number of layers increases ---- the oversquashing phenomenon that occurs when aggregating messages across long paths.

\section{IMPROVE ASSESSMENTS}

\subsection{Activation Functions}
The key to the graph attention network is the introduction of the attention mechanism, by assigning different importance to different neighbor nodes according to the attention coefficient, so that the features of neighbors can be extracted in a focused manner during the aggregation process, thus bringing the improvement of the overall performance. Therefore, the choice of the activation function used in the attention calculation process may impact the performance of the model.

The GAT authors do not provide a theoretical explanation for the choice of the activation function used to calculate the attention coefficient. To explore whether other activation functions can improve the performance of the model as the layers deepen, we use LeakyReLU, Sigmoid, and Tanh on the Cora dataset for the node classification task, respectively, increasing the number of GAT layers and observing the performance changes.

\begin{figure}[htbp]
    \centering
    \includegraphics[scale = 0.6]{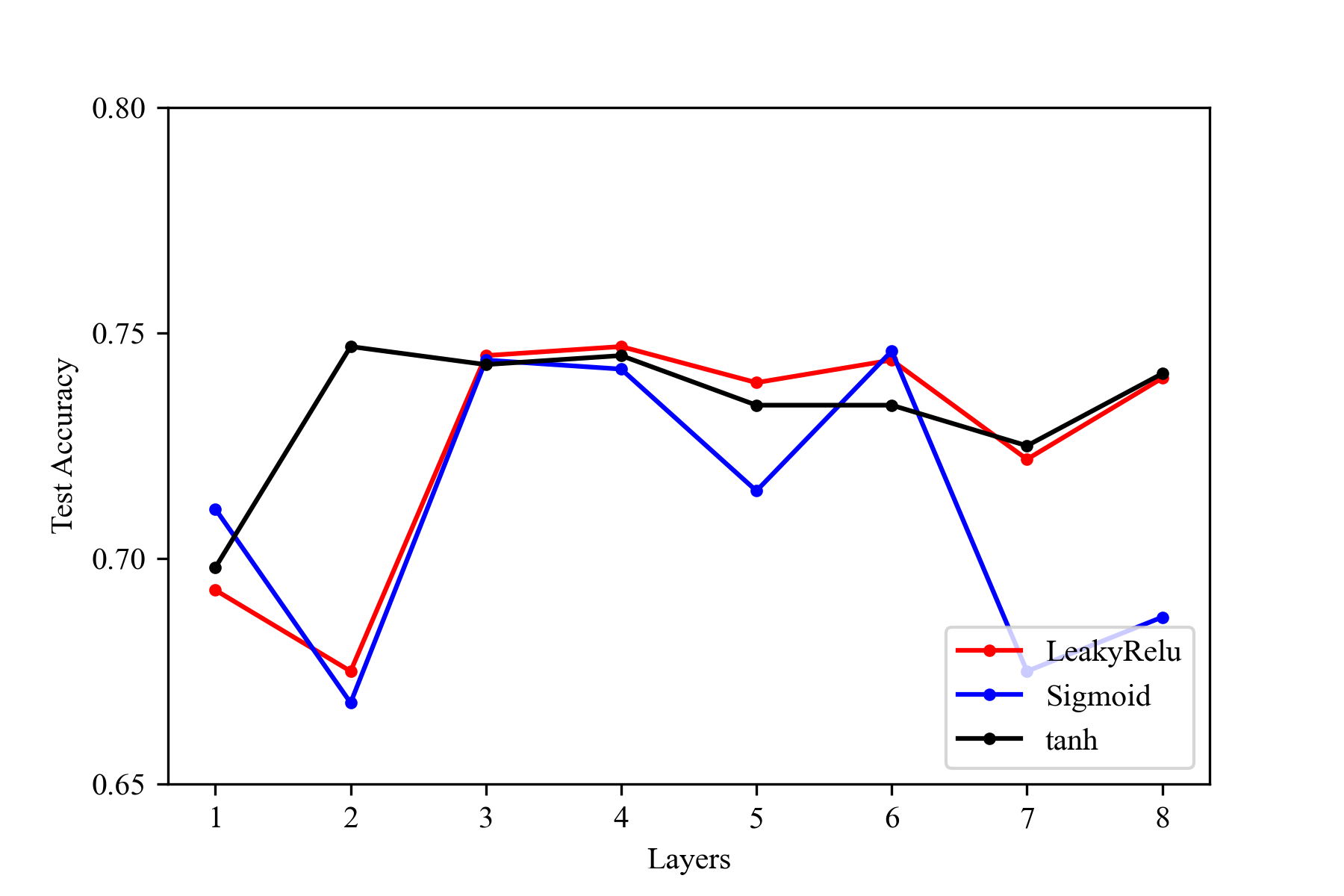}
    \caption{Experimental results of activation functions.}
    \label{fig6}
\end{figure}

The experimental results are shown in the Figure \ref{fig6}. When using the three activation functions, the model accuracy fluctuates substantially and stays around 0.7-0.75 with the increase in the number of layers. It is evident that the choice of the activation function in the attention mechanism is not a relevant factor affecting the performance of GAT with a large number of layers.

\subsection{FA }
Uri AlonEran, Yahav, et al.\cite{DBLP:journals/corr/abs-2006-05205} not only discovered the over-squeezing phenomenon in GNN, but also proposed a simple solution: using fully-adjacent layer (FA).

\begin{figure}[htbp]
\begin{minipage}[t]{0.45\linewidth}
\centering
    \includegraphics[scale = 1]{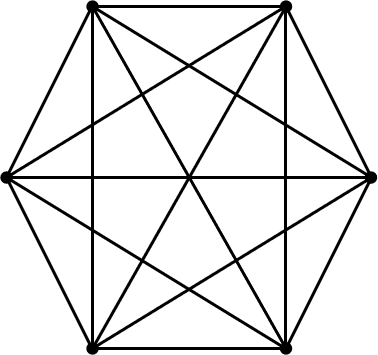}
    \caption{FA Layer}
    \label{fig7}
\end{minipage}%
\begin{minipage}[t]{0.45\linewidth}
\centering
        \includegraphics[scale = 0.6]{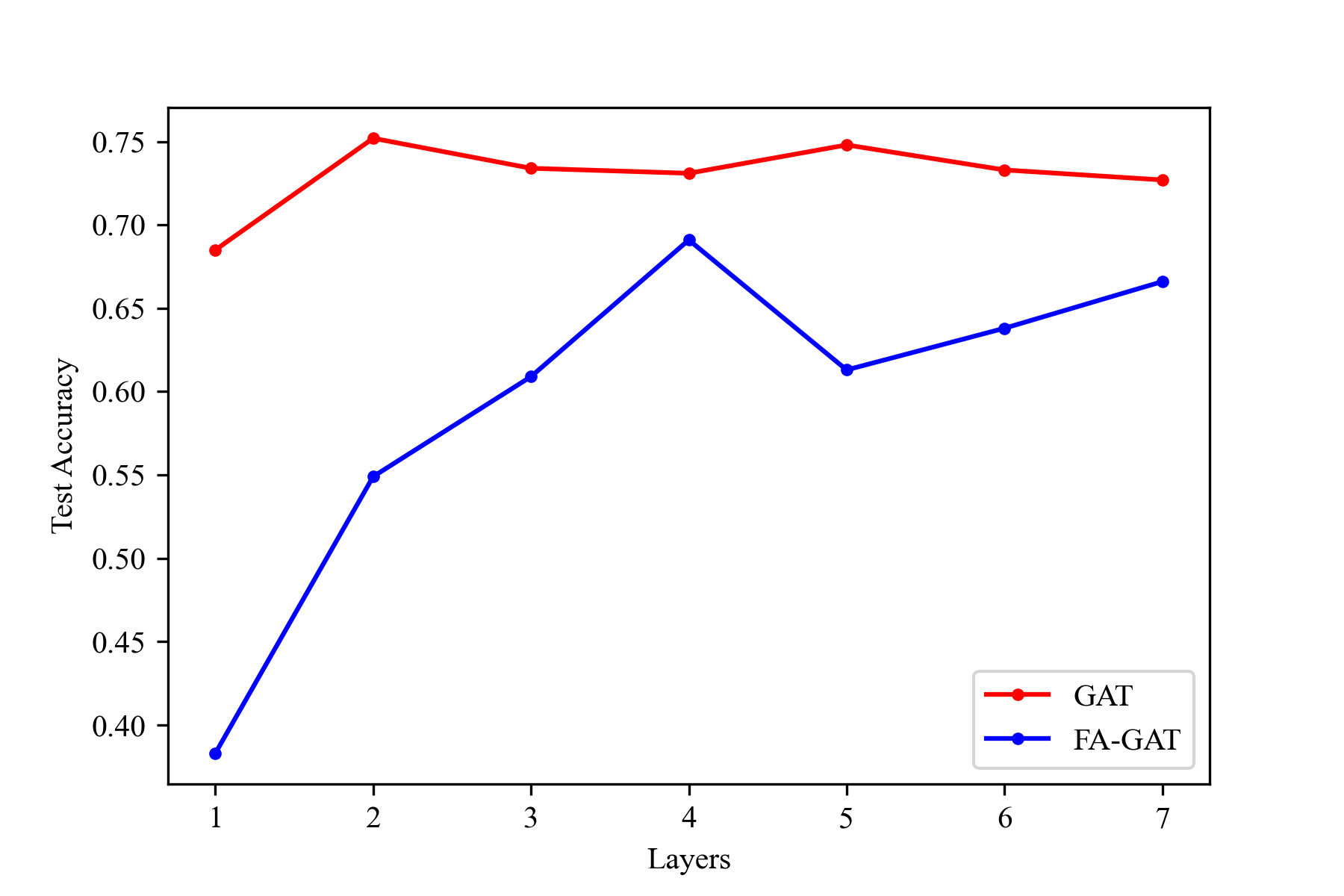}
    \caption{Experimental result of FA.}
    \label{fig8}
\end{minipage}%
\end{figure}

Given a GNN with K layers, adding a FA means modifying the K-th layer to be a fully-adjacent layer(Figure \ref{fig7}). A fully-adjacent layer is a GNN layer in which every pair of nodes is connected. The K - 1 graph layers exploit the graph structure using their original sparse topology, and only the K-th layer is an FA layer that allows the topology-aware node-representations to interact directly and consider nodes beyond their original neighbors. This work argues that this improvement simplifies information flow, prevents over-squashing, and reduces the impact of pre-existing bottlenecks.
We use FA on the GAT model to see if it can alleviate the over-squashing problem of GAT. Considering the large scale when calculating the fully-adjacent layer, we choose the Cora dataset with fewer edges and nodes for experiments. As shown in the Figure \ref{fig8}, the experimental results are not ideal. The accuracy of the GAT model with FA is even substantially lower than that of the original GAT. That is, in some tasks, the method of adding FA is not effective for GAT.
\subsection{Disentangle EP and ET Operation }
Wentao Zhang et al.\cite{DBLP:journals/corr/abs-2108-00955} integrates existing GCN variant models, and conducts extensive experiments based on the idea of decoupling EP and ET. This work finds that the main limitation of deep GCNs is the problem of model degradation introduced by a large number of ET operation.

The work proposes that propagation and transformation operation can be decoupled. At this time, EP and ET are independent of each other, and the number of EP operation can be increased independently to avoid the influence of model degradation introduced by large Dt.

\begin{figure}[htbp]
\centering
    \includegraphics[scale = 0.6]{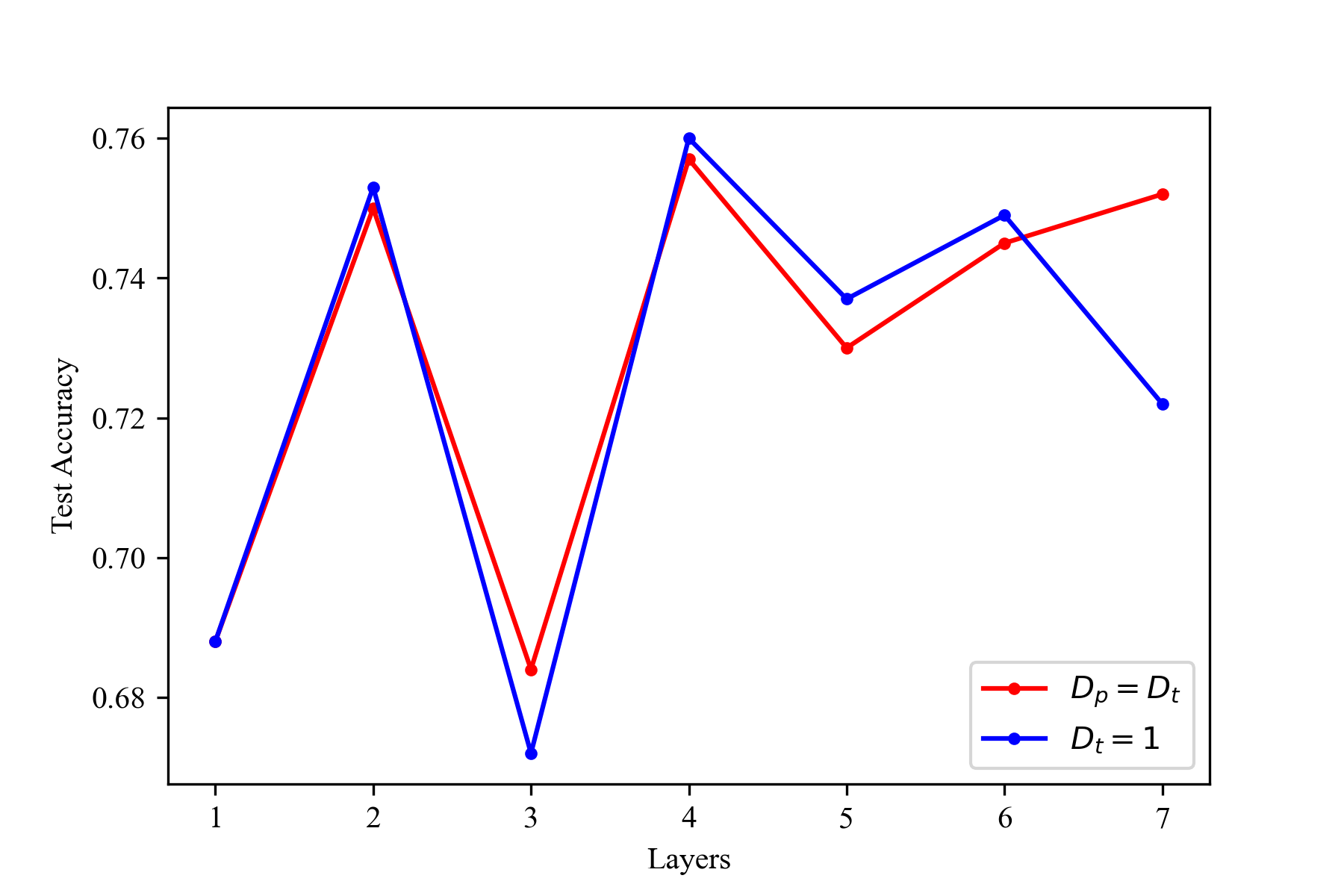}
    \caption{Experimental result of disentangle EP and ET operation}
    \label{fig9}
\end{figure}

The original GAT belongs to the model with entangled ET and EP operation, where the number of ET operation and EP operation is equal, i.e., Dp=Dt. We decouple the GAT and fix its Dt constant to 1. Increase the number of model layers of the original GAT and the decoupled GAT, and compare their test accuracy on the PubMed dataset. Figure \ref{fig9} shows the experimental results.

It is found that when the model becomes deeper, the accuracy of the decoupled GAT does not improve significantly compared with the undecoupled GAT model. The decoupling strategy could not alleviate the possible model degradation problem of GAT, and there is no improvement ability for the GAT.
\subsection{Residual Connection}
When the deep network is propagated forward, the information obtained by the network decreases layer by layer as the network deepens. For such problems, ResNet\cite{DBLP:journals/corr/HeZRS15} proposes an optimization method using residual connections, i.e., the next layer not only includes the new information F(x) after nonlinear transformation of the information of previous layer, but also includes the information x of previous layer. Such treatment effectively solves the problem of information loss, thus achieving higher accuracy in the classification task. This improved perspective fits right into the idea of solving the oversquashing problem.

\begin{figure}[htbp]
\begin{minipage}[t]{0.45\linewidth}
\centering
    \includegraphics[scale = 0.5]{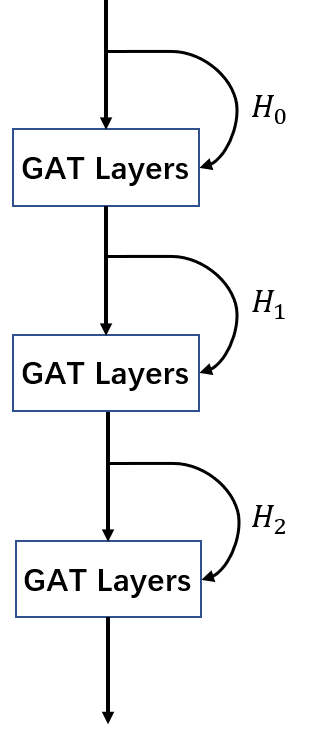}
    \caption{Residual connection.}
    \label{fig10}
\end{minipage}%
\begin{minipage}[t]{0.45\linewidth}
\centering
        \includegraphics[scale = 0.6]{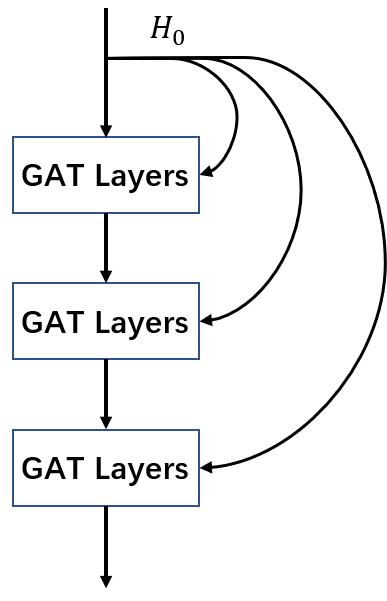}
    \caption{Initial residual connection.}
    \label{fig11}
\end{minipage}%
\end{figure}

We try to use the residual connection on the deep GAT to see if we can get an improvement in accuracy. Referring to the residual connection, we change the output of the GAT layer to a linear superposition of the input and a nonlinear transformation of the input (ie, the output of the original GAT model layer). We set up two modified GATs, one using the normal residual connection (Figure \ref{fig10}); the other using the initial residual connection\cite{DBLP:journals/corr/abs-2007-02133} (Figure \ref{fig11}). Compare the classification task accuracy with the normal GAT on the Pubmed dataset.

\begin{figure}[htbp]
\centering
    \includegraphics[scale = 0.6]{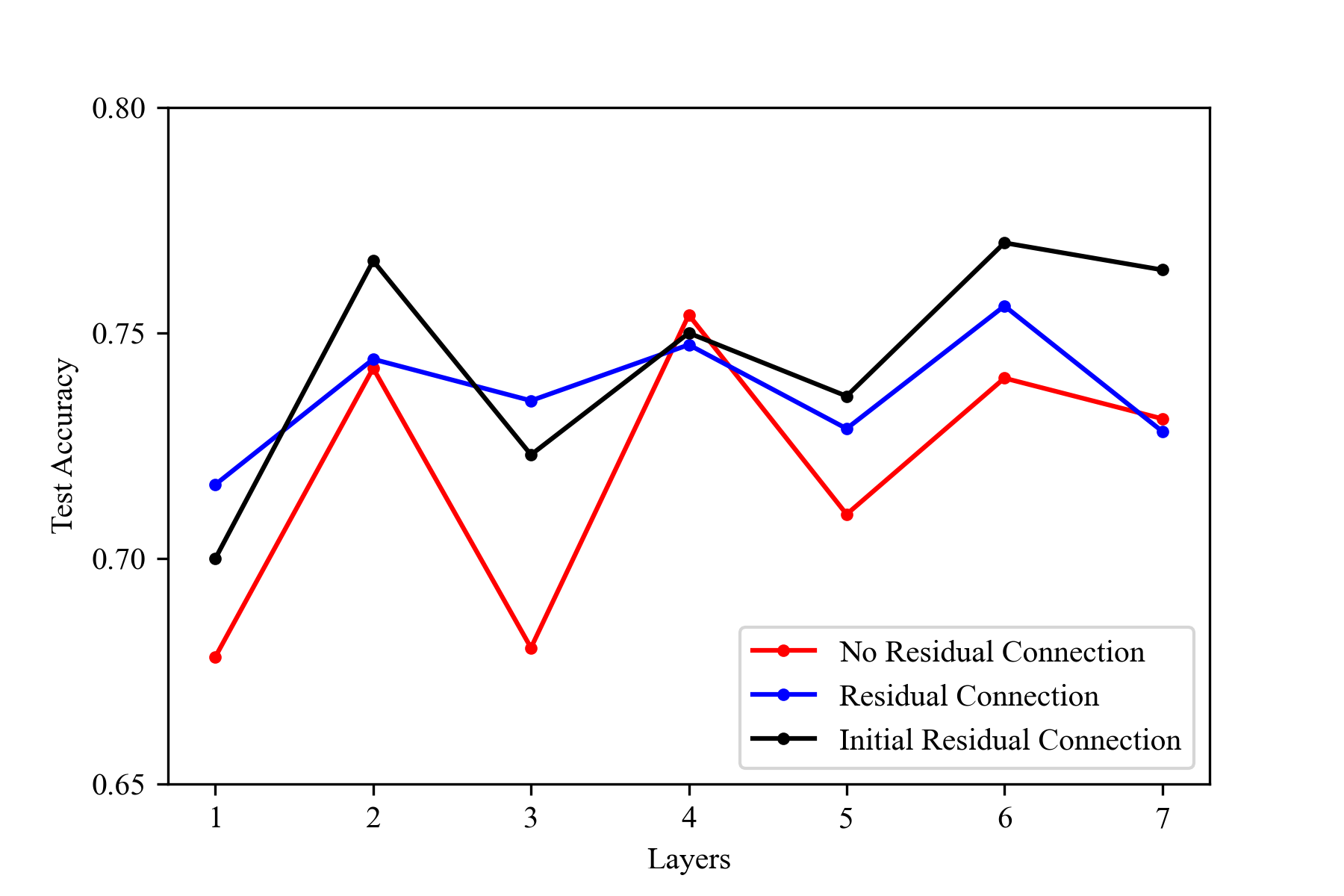}
    \caption{Experimental result of residual connection}
    \label{fig12}
\end{figure}

In Figure \ref{fig12}, the original GAT performance fluctuates substantially with increasing depth, while the performance of the two GATs using residual connections improves steadily, and both increase to some extent over the original GAT. Further, the performance of the initial residual connection is  better than the normal residual connection.

We inferred that the residual connection has an optimization effect on GAT. The explanation we give for this phenomenon is that the residual connection alleviates the over-squashing problem that occurs in GAT during deepening.

\subsection{Analysis}
\begin{figure}[htbp]
\centering
    \includegraphics[scale = 0.5]{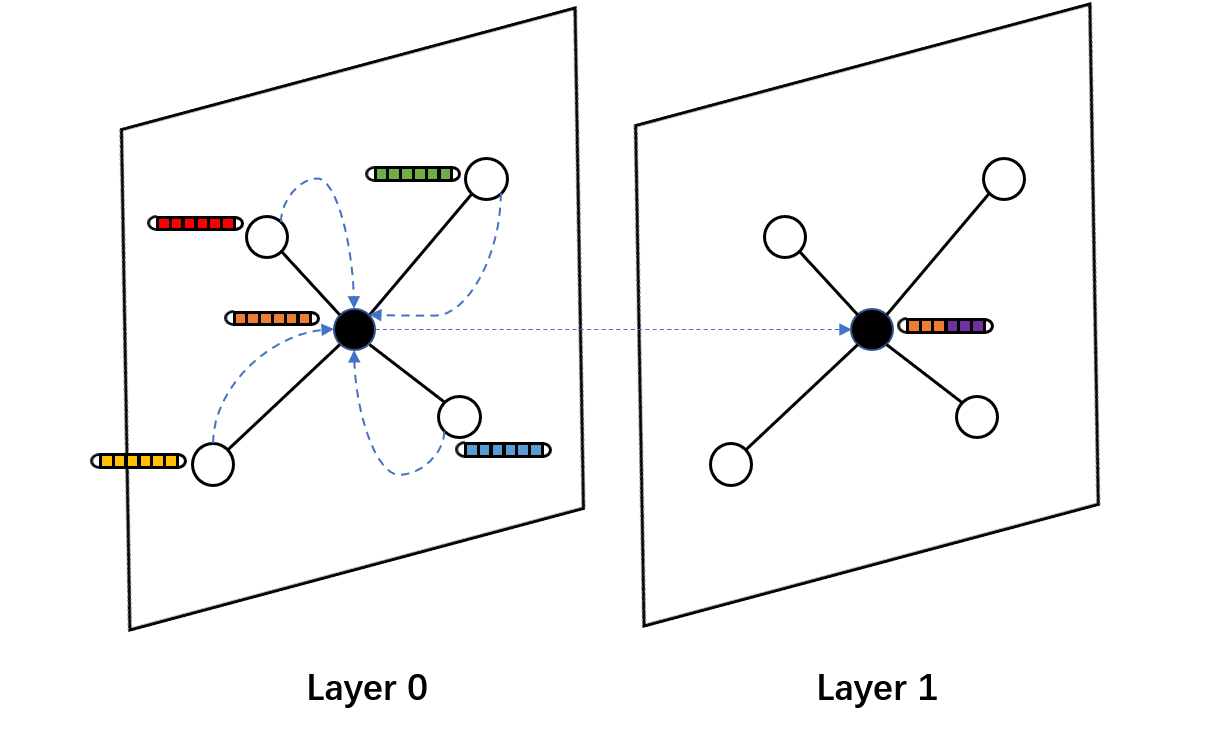}
    \label{fig13}
    \caption{Original information loss}
\end{figure}

As the convolution layer increases, the amount of information received by each node increases exponentially, and thus oversquashing occurs. However, according to our analysis, the oversquashing problem results not only in a large amount of information being compressed in a fixed-length vector, but also an exponential decline of the proportion of original features in the feature representation of each node in the process. (Figure \ref{fig13})

Let $I_o$ denotes the original feature information of node v, and $I_{n_{i}}$ denotes neighbors' feature information where $i\in N$.In the process of calculating the attention coefficient, the information learned through equations 3 and 4 is $\alpha_{i} < I_{o} , I_{n_{i}}>$, and the final node information through the addition of equation 5 is expressed as $<I_{o} ,I_{z} >$, where $I_s$ is the graph structure information learned by node v. When the dimension of the hidden layer remains unchanged, the information capacity of the vector remains unchanged, and the initial feature information of the node is compressed to half of the original, which means that information loss occurs.

$$
\alpha_{i} < I_{o} , I_{n_{i}}>
$$

$$
\alpha_{i} < I_{o} , I_{n_{i}}> + \alpha_{j} < I_{o} , I_{n_{j}}> + ...... + \alpha_{z} < I_{o} , I_{n_{z}}> = <I_{o} ,I_{z} >
$$

GAT uses node features for similarity calculation to represent the importance of node j to node i. When the original feature information in the node representation is overwhelmed by the received information in the process of multiple neighborhood aggregation, the similarity calculation between node features is no longer representative.

\begin{figure}[htbp]
\centering
    \includegraphics[scale = 0.8]{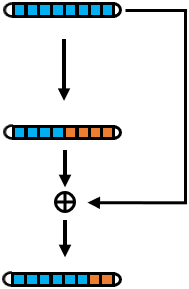}
    \caption{Mitigate original information loss}
    \label{fig14}
\end{figure}

The loss of initial feature information in the nodes is mitigated to some extent by residual connection. The performance improvement of the initial residual connection compared to the ordinary residual connection in the experiment is more significant because the initial residual connection superimposes the initial node feature information on the output of each layer. Compared to the ordinary residual connection, which only mitigates the loss of initial feature information, the initial residual connection can both effectively learn the graph structure information and better maintain the original feature information(Figure \ref{fig14}). In addition, it also alleviates the problem that the computation of attention scores in each GAT layer is affected by neighborhood aggregation.

\section{ADGAT}
 Inspired by the above observations and analysis, we propose the adaptive depth GAT model-ADGAT.

\begin{figure}[htbp]
\centering
    \includegraphics[scale = 0.6]{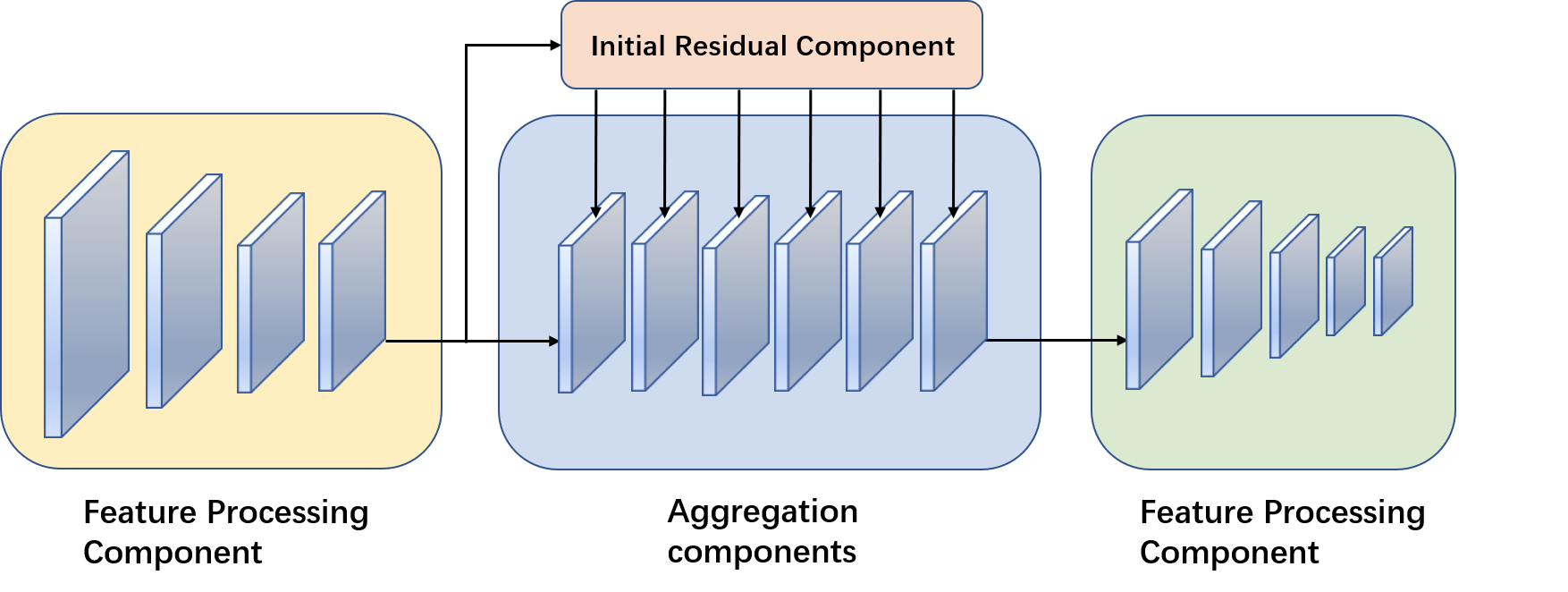}
    \caption{ADGAT}
    \label{fig15}
\end{figure}

Our model is shown in Figure\ref{fig15}. First we use the original feature information processing module, which achieves the effect of processing the initial node information by stacking multiple MLP layers, followed by the initial residual connected convolution module, which adaptively selects the number of convolution layers according to the sparsity of the graph, and learns the graph structure information while maintaining the original feature information. Finally, there is the output feature information processing module, which processes the aggregated node feature information for final output.

 Our model has two outstanding advantages. First, it utilizes an adaptive layer selection mechanism, which can adaptively balance the relationship between the model's receptive field size and the number of layers according to the sparsity of the graph, thus ensuring the receptive field's ability to perceive full graph information while minimizing the degree of oversquashing. Second, it incorporates the initial residual connectivity, thus reducing the loss of original feature information caused by oversquashing.

\subsection{Adaptive Selection of the Number of Layers}
With the increasing popularity of GATs, their adoption extends to domains that require information propagation over longer distances. Such problems require remote interaction and therefore a large number of convolutional layers.

\begin{figure}[htbp]
\centering
    \includegraphics[scale = 0.6]{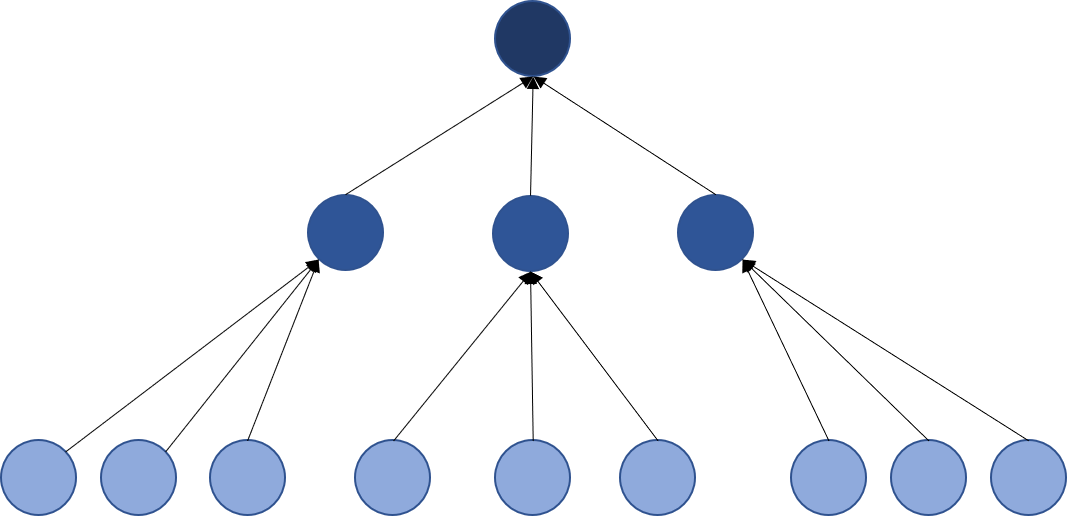}
    \caption{Node tree}
    \label{fig16}
\end{figure}

And for a node to receive K-order neighbor information, the GAT needs to have at least K layers, otherwise, it will suffer from the under-reach problem --- these distant nodes will not know each other at all. However, as the number of layers increases, the number of nodes in each node's receptive field grows exponentially(Figure \ref{fig16}). This leads to overcompression: information from exponentially growing receptive fields is compressed into a fixed-length node vector, and the more layers there are - the more harmful the oversquashing is. Therefore, finding a balance between oversquashing and increases the number of layers is a very important issue.

The focus of this problem is to find the number of GAT layers that will achieve the best possible performance while mitigating the oversquashing problem to the greatest extent possible. We want the number of layers to be the minimum number of layers that the GAT model can have if the receptive field meets the task requirements. This concept is proposed to allow us to obtain the optimal number of design layers to lower the effect of the intensity of oversquashing on the model's effectiveness.

We take the Cora dataset as an example, which has a total of 5429 edges and 2708 nodes in the citation network, where each node is connected to 4 edges on average. The number of connected nodes increases exponentially with each order of receptive  field expansion, using the virtual node v as the average representation of the full graph nodes.

It is worth noting that the receptive field growth index of the node is fixed and computable. As shown in the figure, this index is exactly equal to the average degree of the node. This allows us to have a metric for the degree of information expansion, i.e., to obtain the size of the receptive field as the number of layers increases- the nodes involved in information aggregation.

Intuitively, when the receptive field contains the full graph nodes, i.e., the number of nodes included in the calculated receptive field is approximately equal to the number of full graph nodes, most of the nodes can be aggregated to the full graph information, and the receptive field meets the feature aggregation requirements of the GNN task for the full graph neighbors at this time, so the order at this time is the optimal number of layers.

Based on the above analysis, we propose an adaptive layer calculation method for different data sets.

Given a graph G = (V, E), |V| is the number of nodes, |E| is the number of edges.

$$
\begin{aligned}
q & =\frac{2|E|}{|V|} \\
& \frac{1-q^L}{1-q}=|V| \\
L= & \log _{\frac{2|E|}{|V|}}(1-|V|+2|E|)
\end{aligned}
$$

q is the average number of edges connected at each node, and L is the ideal number of convolution layers designed according to the sparsity of the graph. Since there is an error, the number of layers closest to L at this point is the number of layers adaptively chosen by the model.

\subsection{ADGAT Layer}
From the theoretical analysis in 4.5, it is clear that the initial residual connection can effectively reduce the loss of original feature information due to the oversquashing problem when the number of model layers increases.
Using the initial residual connection, each layer of ADGAT is defined as

\begin{equation}
e\left(\boldsymbol{h}_i, \boldsymbol{h}_j\right)=\operatorname{LeakyReLU}\left(\boldsymbol{a}^{\top} \cdot\left[\boldsymbol{W} \boldsymbol{h}_i \| \boldsymbol{W} \boldsymbol{h}_j\right]\right) \label{e11}
\end{equation}

\begin{equation}
\alpha_{i j}=\operatorname{softmax}_j\left(e\left(\boldsymbol{h}_i, \boldsymbol{h}_j\right)\right) 
=\frac{\exp \left(e\left(\boldsymbol{h}_i, \boldsymbol{h}_j\right)\right)}{\sum_{v_{j^{\prime}} \in \mathcal{N}_i} \exp \left(e\left(\boldsymbol{h}_i, \boldsymbol{h}_{j^{\prime}}\right)\right)}
\label{e12}
\end{equation}

\begin{equation}
\boldsymbol{h}_i^{\prime}=\sigma\left(\sum_{j \in \mathcal{N}_i} \alpha_{i j} \cdot \boldsymbol{W} \boldsymbol{h}_j  +\beta\cdot  \boldsymbol{W^{\prime}}  \boldsymbol{h}_i \right) \label{e13}
\end{equation}

where $\boldsymbol{h}_i$ denotes the representation of the node i, the input to this layer and  $\boldsymbol{h}_i^{\prime}$ is the output of this layer. 
Eq.\ref{e11} calculates the attention score between a pair of nodes i and j, where $\boldsymbol{a} \in \mathbb{R}^{2 d^{\prime}} , \boldsymbol{W} \in \mathbb{R}^{d^{\prime} \times d}
$ are learnable parameters, $\|$ represents the concat operation, and LeakyRelu is an activation function. Eq.\ref{e12} normalizes the attention scores of node i and its neighbors $j \in \mathcal{N}_i$ obtained by Equation 3. Eq.\ref{e13} performs a weighted summation of the neighbor node representations according to the normalized attention scores and initial residual connection. $\beta$ is an artificially set hyperparameter and  $\boldsymbol{W} \in \mathbb{R}^{d^{\prime} \times d}$  is used to align dimensions.

\section{EXPERIMENT}
In this section, we evaluate the performance of ADGAT in a real-world benchmark under semi-supervised learning.

\subsection{Dataset}
Three  citation network datasets (Pubmed\cite{Sen2008CollectiveCI}, Cora\cite{Sen2008CollectiveCI}, and Citeseer\cite{Sen2008CollectiveCI}) are used to benchmark the performance of ADGAT. The statistics of these datasets and data splits could be found in Table \ref{t1}. We use the GAT and ADGAT models to perform node classification tasks on each of the three datasets and record the experimental results.

\subsection{Model and Experiment Settings}

We use GAT as the benchmark model as a comparison to verify the superior performance of our model. The two models with one to eight convolutional layers, ten random seeds randomly selected under different convolutional layers are trained on the train set, and the model that achieves the best results on the validation set is taken to the test set for test and averaged as the final result. We search learning rate within {1, 5}×\{$10-6$, $10^{-5}$, $10^{-4}$, $10^{-3}$, $10^{-2}$\} in all cases, and {0, 1, 2, 5} × \{$10-6$, $10^{-5}$, $10^{-4}$, $10^{-3}$ \} for weight decay rate. For ADGAT, we choose the $\beta$ value among \{0.2, 0.5, 1, 2, 5\}.

\subsection{Comparison With Baseline Model}
The results are shown in the Table \ref{t2}. "Average" indicates the average of the results of the models with different convolutional layers in the data set. It can be seen that ADGAT has achieved a superior performance  in most layers compared to GAT, and  the models with the best results  on each dataset are all ADGAT.  At the same time, the number of layers that ADGAT achieves the best performance on each data set verifies the theoretical content of 5.1.

\begin{table}[htbp]
 \caption{Overview of datasets and task types.}
  \centering
$$
\begin{array}{ccccccc}
\hline \text { Dataset } & \text { \#Nodes } & \text { \#Features } & \text { \#Edges } & \text { \#Classes } & \text { \#Train/Val/Test } & \text { Description } \\
\hline \text { Cora } & 2,708 & 1,433 & 5,429 & 7 & 140 / 500 / 1,000 & \text { citation network } \\
\text { Citeseer } & 3,327 & 3,703 & 4,732 & 6 & 120 / 500 / 1,000 & \text { citation network } \\
\text { Pubmed } & 19,717 & 500 & 44,338 & 3 & 60 / 500 / 1,000 & \text { citation network } \\
\hline
\label{t1}
\end{array}
$$
  \label{tab:table1}
\end{table}

\begin{table}[htbp]
 \caption{Summary of classification accuracy (\%) results among deep GAT-based methods. The value of “Average” shows the average accuracy  between models with 1 and 8 layers.}
  \centering
$$
\begin{array}{ccccccccccc}
\hline \hline & & & & &{\text { Layers }} & & & & & \\
\text { Dataset } & \text { Method } & 1 & 2 & 3 & 4 & 5 & 6 & 7 & 8 & \text { Average } \\
\hline \text { Cora } & \text { GAT } & {68.5} & \mathbf{75.2} & 73.4 & 73.1 & 74.8 & 73.3 & 72.7 & 71.7   & 72.8 \\
& \text { ADGAT } & 68.7 & 74.2 & 74.6 & 75.1 & \mathbf{76.8} & 74.9 & 73.5& 72.5 & 73.8\\
\hline \text { Citeseer } & \text { GAT } & {60.4} & 61.7 & \mathbf{62.2} & 61.9 & 59.7 & 60.5 & 60.9 & 60.2   & 60.9 \\
& \text { ADGAT } & 60.3 & 62.5 & 62.4 & 62.2 & 61.1 & 61.3 & \mathbf{63.5} & 63.4 & 62.0\\
\hline \text { Pubmed } & \text { GAT } & {67.0} & 73.3 & 73.4 & \mathbf{ 76.8} & 75.3 & 73.3 & 74.2 & 72.9   & 73.3 \\
& \text { ADGAT } & 69.9 & 73.7 & 74.9 & 75.4 & 75.9 & 76.2 & \mathbf{77.8} & 76.3 & 75.0 \\
\hline
\label{t2}
\end{array}
$$
  \label{tab:table2}
\end{table}

\section{CONCLUSION}

In this paper, we explore the main factors affecting the performance of GAT when the number of layers is increasing, and find that the initial residual connection can reduce the loss of original feature information caused by oversquashing when the number of layers increases, and explain it theoretically. Based on this, we design the adaptive depth GAT-ADGAT, and verify its superior performance through experiments.

\bibliographystyle{unsrt}  
\bibliography{references}

\begin{thebibliography}{10}

\bibitem{inproceedings}
Marco Gori, Gabriele Monfardini, and Franco Scarselli.
\newblock A new model for earning in raph domains.
\newblock volume~2, pages 729 -- 734 vol. 2, 01 2005.

\bibitem{DBLP:journals/tnn/ScarselliGTHM09}
Franco Scarselli, Marco Gori, Ah~Chung Tsoi, Markus Hagenbuchner, and Gabriele
  Monfardini.
\newblock The graph neural network model.
\newblock {\em {IEEE} Trans. Neural Networks}, 20(1):61--80, 2009.

\bibitem{DBLP:journals/tnn/Micheli09}
Alessio Micheli.
\newblock Neural network for graphs: {A} contextual constructive approach.
\newblock {\em {IEEE} Trans. Neural Networks}, 20(3):498--511, 2009.

\bibitem{DBLP:journals/corr/abs-1902-07243}
Wenqi Fan, Yao Ma, Qing Li, Yuan He, Yihong~Eric Zhao, Jiliang Tang, and Dawei
  Yin.
\newblock Graph neural networks for social recommendation.
\newblock {\em CoRR}, abs/1902.07243, 2019.

\bibitem{DBLP:journals/corr/abs-2110-03987}
Chao Huang, Huance Xu, Yong Xu, Peng Dai, Lianghao Xia, Mengyin Lu, Liefeng Bo,
  Hao Xing, Xiaoping Lai, and Yanfang Ye.
\newblock Knowledge-aware coupled graph neural network for social
  recommendation.
\newblock {\em CoRR}, abs/2110.03987, 2021.

\bibitem{DBLP:journals/corr/MontiBB17}
Federico Monti, Michael~M. Bronstein, and Xavier Bresson.
\newblock Geometric matrix completion with recurrent multi-graph neural
  networks.
\newblock {\em CoRR}, abs/1704.06803, 2017.

\bibitem{DBLP:journals/csur/WuSZXC23}
Shiwen Wu, Fei Sun, Wentao Zhang, Xu~Xie, and Bin Cui.
\newblock Graph neural networks in recommender systems: {A} survey.
\newblock {\em {ACM} Comput. Surv.}, 55(5):97:1--97:37, 2023.

\bibitem{DBLP:journals/kbs/YinLZL19}
Ruiping Yin, Kan Li, Guangquan Zhang, and Jie Lu.
\newblock A deeper graph neural network for recommender systems.
\newblock {\em Knowl. Based Syst.}, 185, 2019.

\bibitem{DBLP:journals/corr/abs-2102-10056}
Yuyang Wang, Jianren Wang, Zhonglin Cao, and Amir~Barati Farimani.
\newblock Molclr: Molecular contrastive learning of representations via graph
  neural networks.
\newblock {\em CoRR}, abs/2102.10056, 2021.

\bibitem{DBLP:journals/jcheminf/JiangWHCLWSCWH21}
Dejun Jiang, Zhenxing Wu, Chang{-}Yu Hsieh, Guangyong Chen, Ben Liao, Zhe Wang,
  Chao Shen, Dong{-}Sheng Cao, Jian Wu, and Tingjun Hou.
\newblock Could graph neural networks learn better molecular representation for
  drug discovery? {A} comparison study of descriptor-based and graph-based
  models.
\newblock {\em J. Cheminformatics}, 13(1):12, 2021.

\bibitem{DBLP:journals/corr/abs-2108-00955}
Wentao Zhang, Zeang Sheng, Yuezihan Jiang, Yikuan Xia, Jun Gao, Zhi Yang, and
  Bin Cui.
\newblock Evaluating deep graph neural networks.
\newblock {\em CoRR}, abs/2108.00955, 2021.

\bibitem{DBLP:journals/corr/abs-2006-05205}
Uri Alon and Eran Yahav.
\newblock On the bottleneck of graph neural networks and its practical
  implications.
\newblock {\em CoRR}, abs/2006.05205, 2020.

\bibitem{DBLP:conf/icml/XuLTSKJ18}
Keyulu Xu, Chengtao Li, Yonglong Tian, Tomohiro Sonobe, Ken{-}ichi
  Kawarabayashi, and Stefanie Jegelka.
\newblock Representation learning on graphs with jumping knowledge networks.
\newblock In Jennifer~G. Dy and Andreas Krause, editors, {\em Proceedings of
  the 35th International Conference on Machine Learning, {ICML} 2018,
  Stockholmsm{\"{a}}ssan, Stockholm, Sweden, July 10-15, 2018}, volume~80 of
  {\em Proceedings of Machine Learning Research}, pages 5449--5458. {PMLR},
  2018.

\bibitem{DBLP:conf/iccv/Li0TG19}
Guohao Li, Matthias M{\"{u}}ller, Ali~K. Thabet, and Bernard Ghanem.
\newblock Deepgcns: Can gcns go as deep as cnns?
\newblock In {\em 2019 {IEEE/CVF} International Conference on Computer Vision,
  {ICCV} 2019, Seoul, Korea (South), October 27 - November 2, 2019}, pages
  9266--9275. {IEEE}, 2019.

\bibitem{inproceedings1}
Johannes Gasteiger, Aleksandar Bojchevski, and Stephan Günnemann.
\newblock Predict then propagate: Graph neural networks meet personalized
  pagerank.
\newblock 02 2019.

\bibitem{DBLP:journals/corr/abs-1902-07153}
Felix Wu, Tianyi Zhang, Amauri H.~Souza Jr., Christopher Fifty, Tao Yu, and
  Kilian~Q. Weinberger.
\newblock Simplifying graph convolutional networks.
\newblock {\em CoRR}, abs/1902.07153, 2019.

\bibitem{DBLP:journals/corr/abs-2004-11198}
Emanuele Rossi, Fabrizio Frasca, Ben Chamberlain, Davide Eynard, Michael~M.
  Bronstein, and Federico Monti.
\newblock {SIGN:} scalable inception graph neural networks.
\newblock {\em CoRR}, abs/2004.11198, 2020.

\bibitem{DBLP:journals/corr/abs-1710-10903}
Petar Velickovic, Guillem Cucurull, Arantxa Casanova, Adriana Romero, Pietro
  Li{\`{o}}, and Yoshua Bengio.
\newblock Graph attention networks.
\newblock {\em CoRR}, abs/1710.10903, 2017.

\bibitem{DBLP:journals/corr/abs-2103-04886}
George Dasoulas, Kevin Scaman, and Aladin Virmaux.
\newblock Lipschitz normalization for self-attention layers with application to
  graph neural networks.
\newblock {\em CoRR}, abs/2103.04886, 2021.

\bibitem{su2023simple}
Guangxin Su, Hanchen Wang, Ying Zhang, Xuemin Lin, and Wenjie Zhang.
\newblock Simple and deep graph attention networks, 2023.

\bibitem{DBLP:journals/corr/abs-1907-10903}
Yu~Rong, Wenbing Huang, Tingyang Xu, and Junzhou Huang.
\newblock The truly deep graph convolutional networks for node classification.
\newblock {\em CoRR}, abs/1907.10903, 2019.

\bibitem{10.1145/3534678.3539445}
Wei Jin, Xiaorui Liu, Yao Ma, Charu Aggarwal, and Jiliang Tang.
\newblock Feature overcorrelation in deep graph neural networks: A new
  perspective.
\newblock In {\em Proceedings of the 28th ACM SIGKDD Conference on Knowledge
  Discovery and Data Mining}, KDD '22, page 709–719, New York, NY, USA, 2022.
  Association for Computing Machinery.

\bibitem{DBLP:journals/corr/abs-2010-02863}
Dominique Beaini, Saro Passaro, Vincent L{\'{e}}tourneau, William~L. Hamilton,
  Gabriele Corso, and Pietro Li{\`{o}}.
\newblock Directional graph networks.
\newblock {\em CoRR}, abs/2010.02863, 2020.

\bibitem{Godwin2021VeryDG}
Jonathan Godwin, Michael Schaarschmidt, Alex Gaunt, Alvaro Sanchez-Gonzalez,
  Yulia Rubanova, Petar Velivckovi'c, James Kirkpatrick, and Peter~W.
  Battaglia.
\newblock Very deep graph neural networks via noise regularisation.
\newblock {\em ArXiv}, abs/2106.07971, 2021.

\bibitem{DBLP:journals/corr/abs-2003-08414}
Yimeng Min, Frederik Wenkel, and Guy Wolf.
\newblock Scattering {GCN:} overcoming oversmoothness in graph convolutional
  networks.
\newblock {\em CoRR}, abs/2003.08414, 2020.

\bibitem{DBLP:journals/corr/abs-2102-06462}
Yujun Yan, Milad Hashemi, Kevin Swersky, Yaoqing Yang, and Danai Koutra.
\newblock Two sides of the same coin: Heterophily and oversmoothing in graph
  convolutional neural networks.
\newblock {\em CoRR}, abs/2102.06462, 2021.

\bibitem{DBLP:journals/corr/abs-1909-12223}
Lingxiao Zhao and Leman Akoglu.
\newblock Pairnorm: Tackling oversmoothing in gnns.
\newblock {\em CoRR}, abs/1909.12223, 2019.

\bibitem{DBLP:conf/nips/DefferrardBV16}
Micha{\"{e}}l Defferrard, Xavier Bresson, and Pierre Vandergheynst.
\newblock Convolutional neural networks on graphs with fast localized spectral
  filtering.
\newblock In Daniel~D. Lee, Masashi Sugiyama, Ulrike von Luxburg, Isabelle
  Guyon, and Roman Garnett, editors, {\em Advances in Neural Information
  Processing Systems 29: Annual Conference on Neural Information Processing
  Systems 2016, December 5-10, 2016, Barcelona, Spain}, pages 3837--3845, 2016.

\bibitem{DBLP:journals/corr/HamiltonYL17}
William~L. Hamilton, Rex Ying, and Jure Leskovec.
\newblock Inductive representation learning on large graphs.
\newblock {\em CoRR}, abs/1706.02216, 2017.

\bibitem{DBLP:journals/corr/abs-2006-13318}
Chen Cai and Yusu Wang.
\newblock A note on over-smoothing for graph neural networks.
\newblock {\em CoRR}, abs/2006.13318, 2020.

\bibitem{Sen2008CollectiveCI}
Prithviraj Sen, Galileo Namata, Mustafa Bilgic, Lise Getoor, Brian Gallagher,
  and Tina Eliassi-Rad.
\newblock Collective classification in network data.
\newblock {\em AI Mag.}, 29:93--106, 2008.

\bibitem{DBLP:journals/corr/abs-2003-13663}
Chaoqi Yang, Ruijie Wang, Shuochao Yao, Shengzhong Liu, and Tarek~F.
  Abdelzaher.
\newblock Revisiting "over-smoothing" in deep gcns.
\newblock {\em CoRR}, abs/2003.13663, 2020.

\bibitem{DBLP:journals/corr/abs-1904-03751}
Guohao Li, Matthias M{\"{u}}ller, Ali~K. Thabet, and Bernard Ghanem.
\newblock Can gcns go as deep as cnns?
\newblock {\em CoRR}, abs/1904.03751, 2019.

\bibitem{DBLP:journals/corr/abs-2006-07107}
Kuangqi Zhou, Yanfei Dong, Wee~Sun Lee, Bryan Hooi, Huan Xu, and Jiashi Feng.
\newblock Effective training strategies for deep graph neural networks.
\newblock {\em CoRR}, abs/2006.07107, 2020.

\bibitem{DBLP:journals/corr/abs-1801-07606}
Qimai Li, Zhichao Han, and Xiao{-}Ming Wu.
\newblock Deeper insights into graph convolutional networks for semi-supervised
  learning.
\newblock {\em CoRR}, abs/1801.07606, 2018.

\bibitem{DBLP:journals/corr/abs-1906-01210}
Xiaotong Zhang, Han Liu, Qimai Li, and Xiao{-}Ming Wu.
\newblock Attributed graph clustering via adaptive graph convolution.
\newblock {\em CoRR}, abs/1906.01210, 2019.

\bibitem{DBLP:journals/corr/HeZRS15}
Kaiming He, Xiangyu Zhang, Shaoqing Ren, and Jian Sun.
\newblock Deep residual learning for image recognition.
\newblock {\em CoRR}, abs/1512.03385, 2015.

\bibitem{DBLP:journals/corr/abs-2007-02133}
Ming Chen, Zhewei Wei, Zengfeng Huang, Bolin Ding, and Yaliang Li.
\newblock Simple and deep graph convolutional networks.
\newblock {\em CoRR}, abs/2007.02133, 2020.

\end{thebibliography}

\end{document}